\documentclass[sigconf,nonacm,screen,9pt,numbers]{acmart}

\AtBeginDocument{%
  }
    
\AtBeginDocument{
    \hypersetup{
        colorlinks=true,
        citecolor=blue,
        linkcolor=blue,
        urlcolor=blue
    }
}

\usepackage{pifont}
\usepackage{graphicx}
\usepackage{subcaption}
\usepackage{multirow} 
\usepackage{algorithm}
\usepackage{algpseudocode} 

\usepackage{makecell}
\usepackage{enumitem}

\usepackage{etoolbox}
\BeforeBeginEnvironment{equation}{\vspace{-5pt}}
\AfterEndEnvironment{equation}{\vspace{-2pt}}

\usepackage{array}
\begin{document}
\vspace{-65pt}
\title{Two Teachers Better Than One: Hardware-Physics Co-Guided Distributed Scientific Machine Learning}

\author{Yuchen Yuan\textsuperscript{*\dag\,1}, Junhuan Yang\textsuperscript{\dag\,1}, Hao Wan\textsuperscript{1}, Yipei Liu\textsuperscript{1}, Hanhan Wu\textsuperscript{1}, Youzuo Lin\textsuperscript{2}, and Lei Yang\textsuperscript{*\,1}}
\affiliation{%
  \institution{\textsuperscript{1}George Mason University, Fairfax, VA, USA \quad \textsuperscript{2}University of North Carolina at Chapel Hill, NC, USA\\[1pt]
  \normalfont\textsuperscript{\dag}These authors contributed equally. \quad \textsuperscript{*}Corresponding authors. \quad \{yyuan21, lyang29\}@gmu.edu}
  \country{}
}

\renewcommand{\shortauthors}{}

\begin{abstract}

Scientific machine learning (SciML) is increasingly applied to in-field processing, controlling, and monitoring; however, wide-area sensing, real-time demands, and strict energy and reliability constraints make centralized SciML implementation impractical. Most SciML models assume raw data aggregation at a central node, incurring prohibitively high communication latency and energy costs; yet, distributing models developed for general-purpose ML often breaks essential physical principles, resulting in degraded performance.
To address these challenges, we introduce EPIC, a hardware- and physics-co-guided distributed SciML framework, using full-waveform inversion (FWI) as a representative task.
EPIC performs lightweight local encoding on end devices and physics-aware decoding at a central node. By transmitting compact latent features rather than high-volume raw data and by using cross-attention to capture inter-receiver wavefield coupling, EPIC significantly reduces communication cost while preserving physical fidelity. 
Evaluated on a distributed testbed with five end devices and one central node, and across 10 datasets from OpenFWI, EPIC reduces latency by 8.9$\times$ and communication energy by 33.8$\times$, while even improving reconstruction fidelity on 8 out of 10 datasets.

\end{abstract}


\maketitle
\vspace{-8pt}
\section{Introduction}

With the rapid advancement of AI for science, scientific machine learning (SciML) is increasingly adopted to support monitoring and decision-making in real-world applications. 
However, when these SciML models transition from laboratory studies to field-scale deployments, they need to process continuous streams of high-resolution sensor data while meeting constraints on communication bandwidth, energy consumption, latency, and reliability. 
These practical requirements challenge many assumptions underlying current SciML workflows and necessitate a rethinking of how to design, deploy, and manage SciML models efficiently in the field.

Most existing SciML studies \cite{SciML1, SciML2, SciML3, SciML4}, whether explicitly or implicitly, follow a centralized design, assuming that raw measurements collected across distributed sensors are transmitted to a central server for inference.
Data communication in such a scheme can easily become a performance bottleneck: high-rate sensor streams quickly saturate the limited bandwidth, and long-distance data transfer introduces substantial latency and energy overhead.
Moreover, these studies make the ideal assumption that all data can reach the central node, and that a single point of failure can compromise the entire model, which is critically vulnerable in remote or harsh environments. As a result, existing centralized SciML approaches are poorly suited for wide-area, real-time scientific deployments.

Inspired by advances in edge computing \cite{edge_computing1, edge_computing2, edge_computing3, edge_computing4}, moving computation closer to data sources offers a promising solution to alleviate the communication bottlenecks.
However, simply applying existing ML designs in a distributed manner is nontrivial: many SciML tasks rely on global physical coupling, and partitioning computation across devices can violate these physical dependencies, severely degrading inference quality.

To better understand the practical needs and challenges of deploying SciML in the field, this paper takes full-waveform inversion (FWI) --- a physics-governed inverse problem --- as a vehicle to examine design choices of different distributed computing paradigms and their limitations.
FWI reconstructs subsurface velocity models by repeatedly simulating seismic wave propagation governed by partial differential equations (PDEs), which was initially used in geoscience \cite{geoscience1, geoscience2}.
In recent years, data-driven FWI has emerged to address the computational inefficiency of physics-driven FWI and has been applied to other acoustic imaging applications, such as ultrasound in healthcare \cite{healthcare1, healthcare2, healthcare3}.
Like other SciML methods, data-driven FWI (e.g., the representative InversionNet \cite{wu2019inversionnet}) assumes a centralized design with ideal execution environments.

In this paper, we evaluate the centralized InversionNet \cite{wu2019inversionnet} on a distributed computing testbed using wireless communication, which is often required because wired infrastructure is unavailable in remote or harsh environments such as deserts or the open ocean.
Results from an emulated 4G environment validate our hypothesis that communication becomes a performance bottleneck, accounting for up to 93\% of total latency. We also assess two off-the-shelf distributed ML approaches, including a Federated Learning-style Approach (FLA) \cite{FLA1} and a Split Learning-style Approach (SLA) \cite{SLA1}, both of which suffer significant performance degradation.
We investigate the root causes and find that the distributed waveforms from different end devices are highly correlated, making FLA ineffective; moreover, these correlations are highly position-dependent and not equally informative across receivers, making SLA ineffective.

To address the challenges posed by both communication bottlenecks and physical violations, we propose EPIC, an \underline{E}dge-compatible and \underline{P}hysics-\underline{I}nformed distributed S\underline{c}iML framework.
The core component in EPIC is a physics-informed distributed neural network, referred to as EPIC-Net, which interoperates with additional system-level deployment and management modules to address practical challenges in the field. 
EPIC-Net decomposes the inversion pipeline into a lightweight local encoding on end devices to extract compact features and a physics-informed decoding for reconstruction at the central node.
As such, the communication bottleneck can be addressed by transmitting compact features (i.e., latent) instead of raw data; moreover, a cross-attention mechanism \cite{Cross_attention} at the central node ensures that feature fusion captures spatial coupling of wavefields across receivers.
\textbf{Main contributions}:
\begin{itemize}[noitemsep,topsep=2pt]
    \item We devise a hardware-physics co-guided distributed SciML model, namely EPIC-Net, addressing communication bottlenecks while adhering to underlying physical principles.
    \item We develop the holistic EPIC framework, enabling efficient deployment and robust run-time of EPIC-Net onto resource-constrained distributed computing infrastructures.
    \item We validate the EPIC framework on a testbed composed of end devices, demonstrating significant reductions in latency and communication cost, while maintaining high inference accuracy and robustness under network variability.
\end{itemize}

EPIC is evaluated on a distributed testbed with five end devices and one central node, using 10 datasets from OpenFWI \cite{deng2022openfwi} for inversion fidelity test.
Results show that EPIC can achieve up to 8.9$\times$ and 33.8$\times$ reductions in latency and communication energy, respectively, compared with a conventional centralized baseline.
Notably, EPIC also improves reconstruction fidelity on 8 out of 10 datasets, with performance on the remaining 2 datasets only marginally lower (by 0.3\% SSIM).
We attribute this unexpected gain to the hardware-physics co-guidance, which regularizes the network toward physically consistent solutions.
\vspace{-10pt}

\section{Challenge and Hardware-Physics Guidance}
\label{sec:Motivation}

\vspace{3pt}
\noindent\textbf{2.1 Challenge: Communication latency can become bottleneck in the centralized in-field scientific machine learning}

\begin{figure}[b]
\centering
\includegraphics[width=\columnwidth]{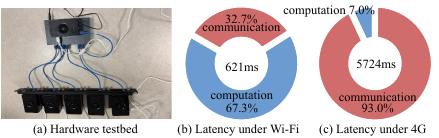}
\caption{In-house testbed and latency breakdown of centralized InversionNet \cite{wu2019inversionnet} across two network settings.
}
\label{fig:latency_breakdown}
\vspace{-10pt}
\end{figure}

We built a hardware testbed of six devices, as shown in Figure \ref{fig:latency_breakdown}(a), including (1) five end devices $(\mathcal{D}_1, \ldots, \mathcal{D}_5)$ that collect input sensor data and (2) one central device ($\mathcal{D}_c$) that generates the final output velocity map.
The testbed can run under Wi-Fi through a gigabit router for communication; it also emulates a high-quality 4G cellular network by limiting the upload bandwidth to 15 Mbps, setting a 50ms latency, and a 0.5\% packet loss rate \cite{4G}.

To identify the performance bottleneck, we measure the proportion of time spent on computation and communication across devices, as reported in 
Figure \ref{fig:latency_breakdown}(b)(c), for real-world Wi-Fi and 4G emulation settings.
By deploying InversionNet \cite{wu2019inversionnet} as a centralized ML system,
raw waveform data are distributively collected from $\mathcal{D}_1$ to $\mathcal{D}_5$, then transmitted to $\mathcal{D}_c$ for the inference.
In both cases, we can observe that the communication latency occupies a large portion of the latency breakdown, i.e., 32.7\% for Wi-Fi and 93\% for 4G emulation, which reveals that the performance bottleneck lies in data communication.
What's worse, in-field implementation often requires real-time performance and has limited energy budgets. However, due to high-volume raw data from sensing (e.g., waveform data in FWI applications), transmitting these data inevitably leads to high communication latency and energy consumption.

\begin{figure}[b]
\centering
\includegraphics[width=\columnwidth]{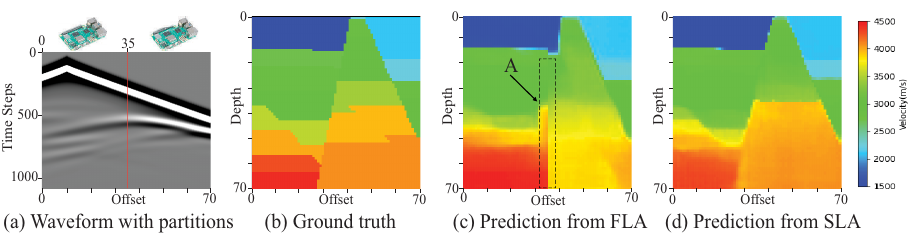}
\caption{Visualization of results from FLA and SLA using 2 distributed end devices, compared with the ground truth.}
\label{fig:artifacts}
\vspace{-10pt}
\end{figure}

\vspace{3pt}
\noindent\textbf{2.2 Hardware Teacher: communication bottleneck can be alleviated by distributed ML, but suffers performance loss}
To alleviate the communication bottleneck, hardware-perspective guidance from edge computing literature directs us to move computation closer to data sources. 
In addition to the centralized ML system (InV) \cite{wu2019inversionnet}, we examine two established distributed ML paradigms:
(1) Federated Learning-style approach (FLA) \cite{FLA1}: processing the distributed raw data on each end device by an entire ML model (i.e., InversionNet in our example) and then combining the partial results (i.e., velocity maps) at the central device. 
(2) Split Learning-style approach (SLA) \cite{SLA1}: extract features from the distributed raw data on each end device by a partial ML model (i.e., the encoder in InversionNet in our example), then combining and further processing these representations at the central device (i.e., by the decoder in InversionNet) to obtain the final result (i.e., the entire velocity map). 
For simplicity, in this section, all distributed systems apply 2 end devices to receive waveform data (see Figure \ref{fig:artifacts}(a)) and 1 central device from the testbed.

\textbf{Communication bottleneck alleviation:} 
Results in Table \ref{tab:method_comparison} demonstrate that both distributed ML paradigms significantly reduce communication costs in the 4G scenario. 
Owing to transmitting compressed representations which are orders of magnitude smaller than raw waveform data, FLA decreases communication time from 5325.52ms to 163.07ms and energy from 2417.786mJ to 74.034mJ, achieving a 96.9\% reduction. Meanwhile, SLA has similar efficiency with 163.13ms latency and 74.061mJ energy consumption. 
Results verify that distributed ML can effectively alleviate the communication bottleneck observed in the centralized implementation.

\begin{table}[t]
\centering
\footnotesize
\captionsetup{font=footnotesize}
\caption{Latency (ms) breakdown and SSIM comparison under 4G scenario}
\setlength{\tabcolsep}{4pt}
\label{tab:method_comparison}
{%
\begin{tabular}
{|c|ccc|cc|cc|}
\hline
{\textbf{Method}} & \textbf{L\textsubscript{edge}} & \textbf{L\textsubscript{comm}} & \textbf{L\textsubscript{central}} & \textbf{L\textsubscript{total}} & \textbf{L\textsubscript{vs. InV}} & {\textbf{SSIM}} & \textbf{$\Delta$\textsubscript{vs. InV}} \\
\hline
InV \cite{wu2019inversionnet} & 0.01 & 5325.52 & 398.42 & 5723.95 & - & 0.85 & - \\
FLA \cite{FLA1} & 241.72 & 163.07 & 0.95 & 405.74 & 14.1$\times$ & 0.76 & -10.58\% \\
SLA \cite{SLA1}& 93.99 & 163.13 & 179.45 & 436.57 & 13.1$\times$ & 0.82 & -3.52\%\\
\hline
\end{tabular}
}
\end{table}

\textbf{Performance degradation:} We have another important observation from the results: compared with the centralized solution (InV), both FLA and SLA suffer different degrees of performance degradation.
From the visualization of the reconstructed velocity map (VM) from FLA in Figure~\ref{fig:artifacts}(c), visible artifacts appear at the boundary of results from two end devices (i.e., bounding box A in the figure). 
Such artifacts can be harmful to domain applications, which can lead to a false prediction of subsurface geologic structures.
Although the solution of SLA in Figure \ref{fig:artifacts}(d) addresses this issue, it still suffers significant performance degradation compared to InV in terms of SSIM scores \cite{SSIM}, as shown in Table \ref{tab:method_comparison}.

We attribute the observed performance degradation to potential violations of fundamental physical principles in distributed strategies.
This perspective is grounded in the broader understanding that scientific ML models must incorporate or preserve physical laws to deliver reliable and high-performing results \cite{physical_laws1, physical_laws2, physical_laws3}, unlike ML for generic datasets (e.g., ImageNet, COCO).

\vspace{3pt}
\noindent\textbf{2.3 Physics Teacher: reveal the root cause of performance degradation, providing guidance on distributed ML designs}

\textbf{Possible root cause 1:} From the perspective of wave physics, the acoustic wave from a source propagates throughout the entire medium. When it reaches a region, the waveform carrying regional information can be transmitted to receivers outside the region, which is determined by the source location, the region's location, and its structure. 
This could be the root cause of performance degradation in solution FLA: the entire ML model on each end device processes waveforms from its corresponding region independently, while ignoring those from others. As a result, each distributed model suffers from information loss, limiting its performance.

\begin{figure}[t]
\centering
\includegraphics[width=\columnwidth]{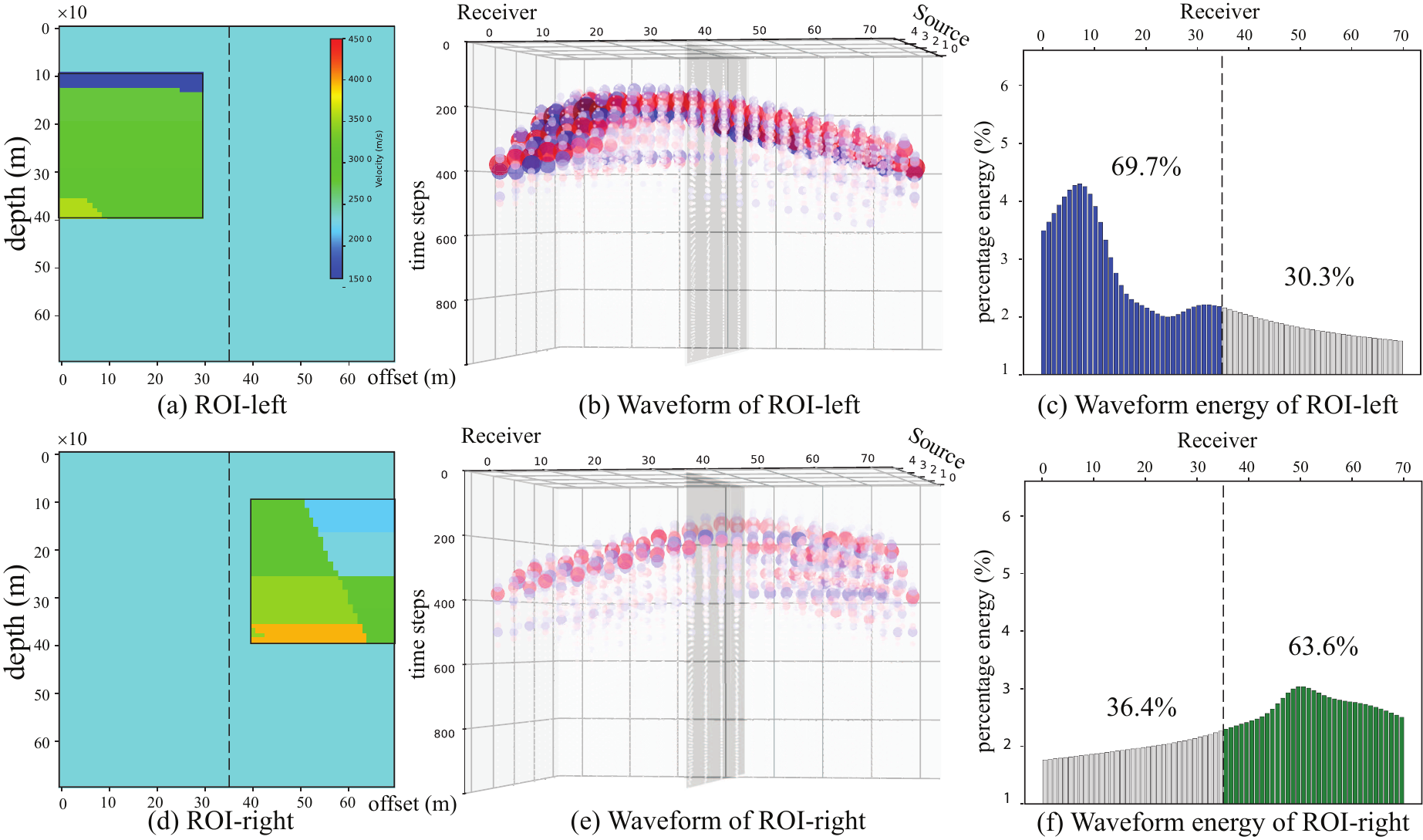}
\caption{Results of exploring two Regions of Interest from the velocity maps (VM) in Figure \ref{fig:artifacts}(b), reporting the corresponding waveforms and energy distributions.}
\label{fig:physics_analysis}
\end{figure}

\textbf{Validation on root cause 1:} 
To validate this hypothesis, we leverage the superposition principle of waves and use Forward Modeling \cite{geoscience1} to capture the corresponding waveforms of two Regions of Interest (ROIs) on velocity maps, shown in Figure \ref{fig:physics_analysis}(a)(d).
To obtain the simulated waveform of ROI-left, we first compute the waveform corresponding to the VM in Figure \ref{fig:physics_analysis}(a), then it is subtracted from the waveform corresponding to VM with a uniform velocity (as background in ROI-left).
We apply the same procedure to ROI-right.
Resultant differential waveforms are shown in Figure \ref{fig:physics_analysis}(b)(e) in 3-D space spanning sources, receivers, and timesteps.

We have a proof-of-concept observation from these visualization results:
even though ROI-Left is only from the left portion of the subsurface, receivers on the right portion still capture substantial signals. ROI-Right exhibits the same phenomenon. 
It validates our hypothesis that FLA suffers performance degradation due to information loss for each distributed model. 
Results of SLA further validate the hypothesis, since the merge of latent representations at the central device exchanges the information from different regions, which provides performance gain over FLA, as shown in Table~\ref{tab:method_comparison}.

\textbf{Possible root cause 2:}
From physics, the strength of reflected waveforms varies spatially: for a given subsurface region, different receivers capture signals with varying amplitudes depending on their positions; similarly, a single receiver records different signal strengths from different regions. This position-dependent behavior suggests that signals from different spatial locations contribute unequally to the reconstruction of various regions in velocity models. 
This could be the root cause of performance degradation in SLA, where all latents are considered equally, ignoring the position-dependent nature of wave propagation. Although coarse-scale structure from waveforms can be reconstructed, it fails to capture fine-scale details (see visualization in Figure \ref{fig:artifacts}(d)), leading to low SSIM.

\textbf{Validation on root cause 2:} 
Figure~\ref{fig:physics_analysis}(c) and (f) quantify this position-dependent signal distribution through energy distribution histograms. For ROI-Left reconstruction, receivers on the left side capture 69.7\% of the total energy; consistently, receivers on the right side capture 63.6\% of energy for ROI-right.
From the distributed ML perspective, these waveforms are processed by different encoders in SLA depending on receiver locations, which generates latent representations and should be weighted when contributing to the later reconstruction process in decoder.
\vspace{-5pt}
\section{EPIC Framework}
\label{sec:EPIC}

\begin{figure}[t]
\centering
\includegraphics[width=\columnwidth]{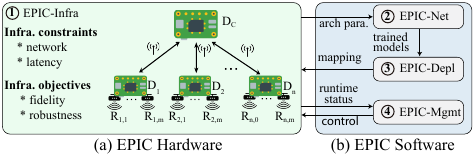}
\caption{Overview of the proposed EPIC framework.}
\label{fig:infra}
\end{figure}

Figure \ref{fig:infra} illustrates the overview of the proposed \underline{E}dge-compatible and \underline{P}hysics-\underline{I}nformed distributed S\underline{c}iML framework, namely EPIC, for high fidelity, low latency, and robust data-driven in-field FWI.
It contains four components: {\Large{\ding{172}}} EPIC-Infra, {\Large{\ding{173}}} EPIC-Net, {\Large{\ding{174}}} EPIC-Depl, and {\Large{\ding{175}}} EPIC-Mgmt.
Specifically, {\Large{\ding{172}}} EPIC-Infra formulates the distributed computing infrastructure with constraints and objectives; {\Large{\ding{173}}} EPIC-Net takes the infrastructure parameters (e.g., the number of end nodes) from EPIC-Infra to construct a distributed neural network architecture; the trained model from EPIC-Net will be passed to {\Large{\ding{174}}} EPIC-Depl for the deployment onto EPIC-Infra; finally, at run-time, {\Large{\ding{175}}} EPIC-Mgmt will monitor and control the system when detecting possible violation of real-time requirements.

\vspace{3pt}
\noindent{\large\ding{192}} \textbf{Resource-constrained and multi-objective EPIC-Infra}

The infrastructure involves a distributed sensing system where seismic data acquisition and processing are spatially partitioned across multiple devices. 
We denote $\mathcal{D} = \{\mathcal{D}_1, \mathcal{D}_2, \ldots, \mathcal{D}_n\}$ as a set of $n$ end devices, and $\mathcal{D}_c$ is the central device.
Each end device $\mathcal{D}_i$ manages a set of receivers $\mathcal{R}_i = \{\,\mathcal{R}_{i,1},\ldots,\mathcal{R}_{i,m}\,\}$ that capture seismic waveforms from its designated spatial region.
In addition to the fundamental sensing and computing components, we also have two constraints in EPIC-Infra.
First, a network $\mathcal{N}=\langle b, l, p\rangle$ is defined as a tuple, where $b$ denotes upload bandwidth, $l$ represents network delay, and $p$ indicates packet loss rate.
Second, scientific applications commonly have a timing constraint $T$.

Based on EPIC-Infra, we can formulate a distributed FWI system.
Taking centralized InV (see Section \ref{sec:Motivation}) as an example, each end device $\mathcal{D}_i \in \mathcal{D}$ will transmit all raw waveform data from all receivers $\forall \mathcal{R}_{i,j} \in \mathcal{R}_i$ to the central device $\mathcal{D}_c$.
A trained InV model is deployed on $\mathcal{D}_c$ for FWI task.
As shown in Table \ref{tab:method_comparison}, high communication latency can lead to violation of timing constraint $T$.

In this work, our objective is not only to perform the FWI task on EPIC-Infra to satisfy constraint $T$, but also to achieve high-fidelity results and provide a robust computation for possible network failure.
We formally define the problem:
Given an EPIC-Infra with central device $\mathcal{D}_c$, end devices $\mathcal{D}$, receiver $\mathcal{R}_i$ associated to end device $\mathcal{D}_i \in \mathcal{D}$, a network $\mathcal{N}$, and a timing constraint $T$, we aim to:
\begin{itemize}[noitemsep,topsep=2pt]
    \item Design a distributed neural network $NN$,
    \item Deploy $NN$ to $\mathcal{D}_c$ and $\mathcal{D}$,  
    \item Manage executions of $NN$ on $\mathcal{D}_c$ and $\mathcal{D}$ via network $\mathcal{N}$,
\end{itemize}
such that the system can tolerate a subset of end devices encountering networking failure, while satisfying the timing constraint $T$ with the maximized fidelity measured by SSIM.

\begin{figure}[t]
\centering
\includegraphics[width=\columnwidth]{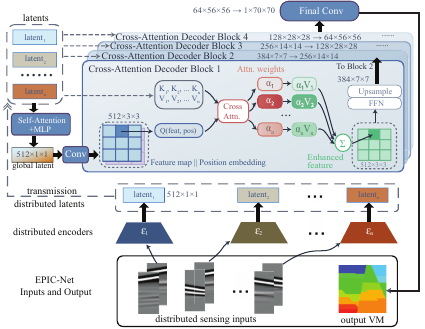}
\caption{Overview of EPIC-Net.}
\label{fig:EPIC_architecture}
\end{figure}

\vspace{3pt}
\noindent{\large\ding{193}} \textbf{Hardware-physics co-guided EPIC-Net}

The key to achieving our ultimate goal is the design of a distributed neural network.
As shown in the motivation (Section \ref{sec:Motivation}), a physics-agnostic neural network design (i.e., FLA and SLA) can achieve high timing performance, but sacrifices the fidelity, which is one of the most important metrics for scientific applications, such as the FWI task.
In this module, we propose a hardware-physics co-guided neural network, namely EPIC-Net, as shown in Figure \ref{fig:EPIC_architecture}. 

Inspired by the performance degradation analysis in Section \ref{sec:Motivation}, EPIC-Net follows SLA's distributed encoder design and integrates physical guidance in designing the centralized decoder.
As shown in Figure~\ref{fig:EPIC_architecture}, there are three major components in EPIC-Net: (1) Distributed Encoding, (2) Self-Attention Fusion, and (3) Position-Aware Cross-Attention Decoder.

\textbf{Distributed Encoding.} According to the EPIC-Infra with $|\mathcal{D}|=n$ end devices $\mathcal{D}$, there will be $n$ {distributed encoders} $\{\varepsilon_1, \varepsilon_2, \ldots, \varepsilon_n\}$.
Each encoder $\varepsilon_i$ on device $\mathcal{D}_i$ processes the waveforms from the associated receivers $\mathcal{R}_i$ through convolutional blocks with progressive downsampling, producing a compact latent representation.
This strategy enables parallel encoding of waveforms into compact latent representations $\mathcal{L}=\{\text{latent}_1, \text{latent}_2, \ldots, \text{latent}_n\}$ on resource-constrained end devices, where $\text{latent}_i \in \mathbb{R}^{512 \times 1 \times 1}$.
The latent will then be transmitted to the central device for reconstruction via the decoding process.
In this way, we can significantly reduce communication cost, compared with transmitting raw waveforms.

\textbf{Self-Attention Fusion.} After receiving $n$ latents from distributed encoders,
EPIC-Net applies a Self-Attention module \cite{Self_attention} that aggregates information from these latents to generate a global latent $gl$.
In this way, we can avoid the performance degradation caused by root cause 1 in Section \ref{sec:Motivation} (i.e., FLA's information loss).

\textbf{Position-Aware Cross-Attention Decoder.} The last component in EPIC-Net is to prevent performance degradation by root cause 2 in Section \ref{sec:Motivation}.
Instead of using the global latent $gl$ only, EPIC-Net leverages $\mathcal{L}$ to implement a position-aware Cross-Attention \cite{Cross_attention}, which will adaptively weight distributed latents based on spatial location.
Specifically, for decoder features at a given resolution, we maintain learnable position embeddings for $70 \times 70$ output space and bilinearly interpolate them to match each decoder resolution. The position-aware query is computed by concatenating decoder features with interpolated position embeddings: $Q(F, E_{\text{pos}}) = \text{Linear}([F; E_{\text{pos}}])$, where $\text{Linear}(\cdot)$ denotes a learnable linear projection. Keys and values are derived from distributed encoder latents: $[K_i, V_i] = \text{Linear}(\text{latent}_i)$. The cross-attention then computes attention weights $\alpha_i$ for each local feature:
\begin{equation}
\small
\alpha_i = \text{softmax}\left(\frac{QK_i^T}{\sqrt{d_k}}\right), \quad \text{Output} = \sum_{i=1}^{n} \alpha_i V_i
\end{equation}
where $d_k$ is the dimension of the key vectors.
The design enables the decoder to weight distributed latents based on their spatial correspondence to the reconstructing region.

\vspace{3pt}
\noindent{\large\ding{194}} \textbf{Automated model-to-device module: EPIC-Depl}

With the neural network architecture defined in EPIC-Net and the hardware infrastructure specified in EPIC-Infra, EPIC-Depl provides an automated pipeline to map trained models onto distributed devices.
Each encoder $\varepsilon_i$ is assigned to end device $\mathcal{D}_i$, where it processes waveforms from receivers $\mathcal{R}_i = \{\mathcal{R}_{i,1}, \ldots, \mathcal{R}_{i,m}\}$, while the decoder runs on central device $\mathcal{D}_c$ for global reconstruction. EPIC-Depl automates: (1) environment configuration on all devices, (2) model weight distribution to corresponding devices, and (3) communication channel establishment through network $\mathcal{N}$. This enables rapid deployment of FWI systems in field environments.

As for the communication component, EPIC-Depl establishes connections between end devices and the central node using an out-of-order packet-handling mechanism with a hash-map buffer, allowing the central node to process samples as soon as $n$ latents from end devices are received.

\vspace{3pt}
\noindent{\large\ding{195}} \textbf{Latency-aware run-time manager: EPIC-Mgmt}

At runtime, end devices may experience network delays or temporary failures, which can cause data to be blocked in the hash map buffer and violate the timing constraint $T$ required by EPIC-Infra. 
We propose a run-time manager, namely EPIC-Mgmt, to monitor transmission status from each end device $\mathcal{D}_i$ and release the pipeline block from the hash-map buffer once their delays exceed the timing constraint $T-T_d$, where $T_d$ is the latency of executing the decoder in EPIC-Net on the central device, which can be obtained by EPIC-Mgmt through a one-time profiling before runtime.

Algorithm~\ref{alg:epic_mgmt} provides the workflow of EPIC-Mgmt, which includes two phases: \textbf{Phase 1}, in which each end device performs encoding and transmits the latent to the central device with a timing threshold $T-T_d$.
\textbf{Phase 2}, the central decoder performs adaptive reconstruction only using successfully received latents within $T-T_d$ time frame. 
Benefiting from the 
position-aware cross-attention mechanism, the decoder can automatically redistribute attention weights when some latents are missing, enabling stable reconstruction even under partial device failures.

\begin{algorithm}[t]
\small 
\caption{EPIC-Mgmt Runtime Control}
\label{alg:epic_mgmt}
\begin{algorithmic}[1]
\Require Waveform splits $\{\mathcal{D}_1, \ldots, \mathcal{D}_n\}$, timeout threshold $T-T_d$
\Ensure Velocity map $\hat{\mathbf{V}}$
\State \textbf{// Phase 1: Edge encoding with timeout monitoring}
\For{each end device $\mathcal{D}_i$ \textbf{in parallel}}
    \State $\text{latent}_i \leftarrow \varepsilon_i(\mathcal{D}_i)$ 
    \State \textbf{if} timeout \textbf{then} 
    \State \ \ \ \ trigger EPIC-Mgmt to release sync block at central node's buffer
\EndFor
\State \textbf{// Phase 2: Adaptive reconstruction at central device}
\State Collect successfully received $\{\text{latent}_i\}$ within $T-T_d$
\State $\hat{\mathbf{V}} \leftarrow \text{Decoder}(\{\text{latent}_i\})$
\end{algorithmic}
\end{algorithm}

\section{Experiment}

\begin{figure}[t]
\centering
\includegraphics[width=\columnwidth]{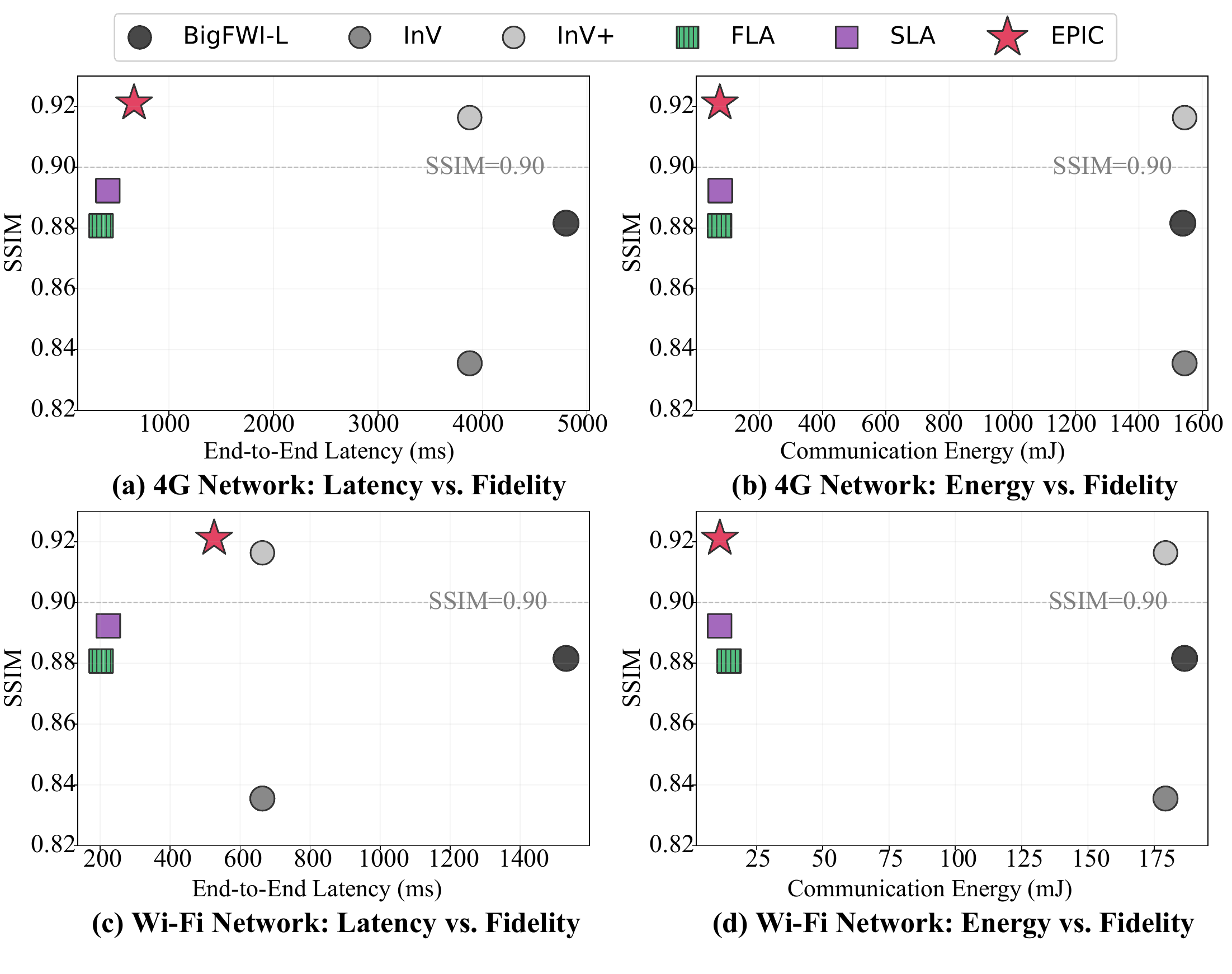}
\caption{Trade-off on communication efficiency and fidelity.}
\label{fig:Latency}
\end{figure}

\vspace{5pt}
\noindent\textbf{4.1 Experimental Setup}
\vspace{3pt}

We use the distributed testbed in Figure \ref{fig:latency_breakdown}(a) to evaluate the proposed EPIC framework.
The testbed is composed of six devices: five Raspberry Pi 5 units (each equipped with 16GB RAM and 128GB storage) serving as distributed end devices, and one serving as the central node. The testbed operates under two network conditions: (1) a real-world Wi-Fi environment through a gigabit router, and (2) an emulated 4G cellular network with 15Mbps upload bandwidth, 50ms latency, and 0.5\% packet loss rate.
Latency and energy consumption are measured by deploying pre-trained models on these devices for inference under both network conditions.

For the evaluation of EPIC-Net, we apply 
OpenFWI dataset~\cite{deng2022openfwi}, which is composed of three families of geophysical subsurface:
(1) \textit{Vel Family} with four datasets on flat/curved velocity variations (denoted as FlatVel-A, FlatVel-B, CurveVel-A, CurveVel-B), 
(2) \textit{Fault Family} with four datasets on geological faults and discontinuities (denoted as FlatFault-A, FlatFault-B, CurveFault-A, CurveFault-B),
and (3) \textit{Style Family} with two datasets on complex patterns from style transfer (denoted as Style-A and Style-B).
Here, each family includes A/B variants with increasing difficulty. 
All samples share consistent dimensions: seismic waveforms 
[5, 1000, 70] (5 sources, 70 receivers, 1000 timesteps) 
paired with velocity maps [70, 70] spanning 1500-4500 m/s.

We compare EPIC-Net against multiple baseline approaches. BigFWI-L~\cite{jin2024empirical} represents the state-of-the-art centralized FWI model. InversionNet (InV)~\cite{wu2019inversionnet} results are directly cited from the OpenFWI benchmark paper, where the model was trained for 120 epochs. To provide a fair comparison under our training protocol, we also report \textbf{InV$+$} results by retraining InversionNet in our environment for 400 epochs with identical hyperparameters for EPIC-Net. Additionally, we compare against two distributed ML paradigms analyzed in Section~2.2: FLA~\cite{FLA1} and SLA~\cite{SLA1}.
All models are trained on NVIDIA A100 GPUs with 40GB memory using the AdamW optimizer~\cite{loshchilov2017decoupled} with an initial learning rate of $1 \times 10^{-3}$. 
We employ the same loss function with equal weights for MAE and MSE terms. 
Training is conducted with a batch size of 32 for 400 epochs.

\vspace{5pt}
\noindent\textbf{4.2 Evaluation of EPIC Framework}
\vspace{3pt}

\textbf{EPIC makes a better trade-off on HW/SW performance.}
Figure~\ref{fig:Latency} demonstrates the trade-off between average software performance (SSIM score) and hardware performance (both latency and energy consumption) across all ten datasets.
We applied two communication modes, emulated 4G and the real-world Wi-Fi network, for testing.
The ideal result lies in the top-left corner for high reconstruction fidelity and low communication cost.

From Figure \ref{fig:Latency},
EPIC consistently achieves the highest SSIM, compared with the centralized approaches.
Moreover, EPIC-Net consumes less energy than distributed methods and an order of magnitude less than centralized approaches.
In other words, EPIC dominates all other solutions in the SSIM-energy trade-offs, as shown in Figure \ref{fig:Latency}(b)(d).
In terms of latency, EPIC reduces end-to-end latency by several-fold compared with all centralized baselines; however, it has higher latency compared with the other two distributed approaches.
This is because the cross-attention brings computation overhead to the central node.
In our testbed, we use the same edge device as the central node; this latency gap will be closed if we use a server-grade computer.
These results demonstrate that EPIC-Net provides superior performance across all deployment conditions with the highest reconstruction accuracy while also achieving the lowest communication cost, validating its effectiveness as a system–physics co-designed distributed solution.

\begin{figure}[t]
\centering
\includegraphics[width=\columnwidth]{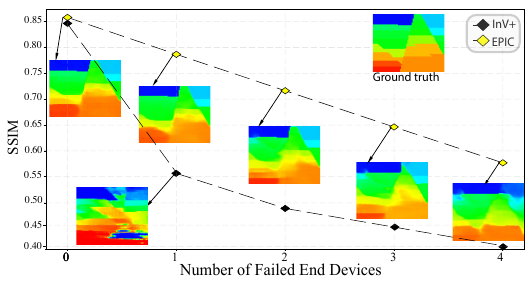}
\caption{Robustness evaluation when communication failures occur with visualizations on results.}
\label{fig:robust}
\end{figure}
\textbf{EPIC provides high robustness.}
A practical requirement in field deployments is robustness to missing or delayed edge nodes.
Figure~\ref{fig:robust} evaluates the robustness of EPIC against the centralized approach on the FlatFault-B dataset by simulating the loss of 0 to 4 out of 5 edge nodes.
From the results, we can see that the performance of InV$+$ degrades dramatically to 0.56, even with only one node missing. 
In contrast, EPIC's average SSIM decreases from 0.86 (no dropout) to 0.79, 0.72, and 0.65, respectively, when 1, 2, and 3 nodes are unavailable.
We also show the visualization of the reconstructed subsurfaces in different situations.
Compared with the ground-truth, the results from EPIC, even with four missing latents, can still extract the layer and fault features; on the other hand, the visualization results of centralized approaches are severely distorted even for only one node missing. 
The robustness of EPIC is from the collaborative effects of EPIC-Net and EPIC-Mgmt.
Overall, EPIC enables continuous operation under disruptions by avoiding global stalls and meeting the deadline~$T$, a key requirement for real-time monitoring in distributed scientific systems.

\vspace{5pt}
\noindent\textbf{4.3 Evaluation of EPIC-Net}
\vspace{3pt}

\textbf{Hardware-physics co-guidance boosts SW performance.}
Table~\ref{tab:main_results} compares EPIC-Net against state-of-the-art baselines across all ten OpenFWI benchmark datasets. 
EPIC-Net achieves the best performance on six datasets and the second-best on three of the remaining four, demonstrating competitive reconstruction quality. 
Notably, compared to the centralized baseline InV+, EPIC-Net improves SSIM scores on 8 out of 10 datasets. 
Compared to distributed baselines, FLA and SLA, the advantages are particularly pronounced on B-variant datasets, which feature more complex geological structures than their A-variant counterparts. 
These results demonstrate that EPIC-Net can not only avoid performance degradation from distribution but also can surpass centralized methods across datasets with diverse geological conditions, showcasing the effectiveness of hardware-physics co-guided design for distributed SciML.

\begin{table}[t]
\centering
\small
\caption{Comparison of EPIC-Net with competitors on OpenFWI datasets. Best results in \textbf{bold}, second best \underline{underlined}.}
\label{tab:main_results}
\resizebox{\columnwidth}{!}{%
\begin{tabular}{lcccccc}
\hline
\textbf{Dataset} & \textbf{BigFWI-L} & \textbf{InV} & \textbf{InV+} & \textbf{FLA} & \textbf{SLA} & \textbf{EPIC-Net} \\
\hline
FlatVel-A & 0.9965 & 0.9895 & \underline{0.9976} & 0.9938 & 0.9944 & \textbf{0.9987} \\
FlatVel-B & 0.9756 & 0.9461 & \underline{0.9817} & 0.9750 & 0.9784 & \textbf{0.9872} \\
CurveVel-A & \textbf{0.9199} & 0.8223 & 0.9153 & 0.8827 & 0.8855 & \underline{0.9198} \\
CurveVel-B & 0.8134 & 0.6727 & \underline{0.8225} & 0.7621 & 0.7759 & \textbf{0.8351} \\
\hline
FlatFault-A & \textbf{0.9918} & 0.9798 & 0.9872 & 0.9719 & 0.9824 & \underline{0.9882} \\
FlatFault-B & 0.8033 & 0.7208 & \textbf{0.8621} & 0.7964 & 0.8178 & \underline{0.8601} \\
CurveFault-A & \textbf{0.9801} & 0.9566 & \underline{0.9719} & 0.9457 & 0.9608 & 0.9687 \\
CurveFault-B & 0.6790 & 0.6163 & \underline{0.7586} & 0.6883 & 0.7258 & \textbf{0.7604} \\
\hline
Style-A & 0.9136 & 0.8910 & 0.9731 & 0.9507 & \underline{0.9732} & \textbf{0.9782}\\
Style-B & 0.7429 & 0.7599 & \underline{0.8935} & 0.8410 & 0.8290 & \textbf{0.9147} \\
\hline
\end{tabular}%
}
\end{table}

\begin{figure}[t]
\centering
\vspace{6pt}
\includegraphics[width=\columnwidth]{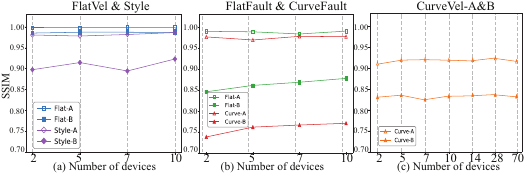}
\caption{The scalability test results of EPIC-Net in terms of the end device numbers.}
\label{fig:scalability}
\end{figure}

\textbf{Scalability.}
A key property of EPIC is its scalability with respect to the number of end devices.
Figure~\ref{fig:scalability}(a)(b) evaluates EPIC-Net under different configurations, ranging from 2 to 10 end devices for all datasets. Across all cases, SSIM remains highly stable with only minor fluctuations.
Figure \ref{fig:scalability}(c) further extends the scale to 70, indicating each receiver is equipped with an end device; it slightly drops compared with the configuration with 28 end devices.
These results show that EPIC retains good scalability on distributed EPIC-Infra.

\begin{figure}[!t]
\centering
\includegraphics[width=\columnwidth]{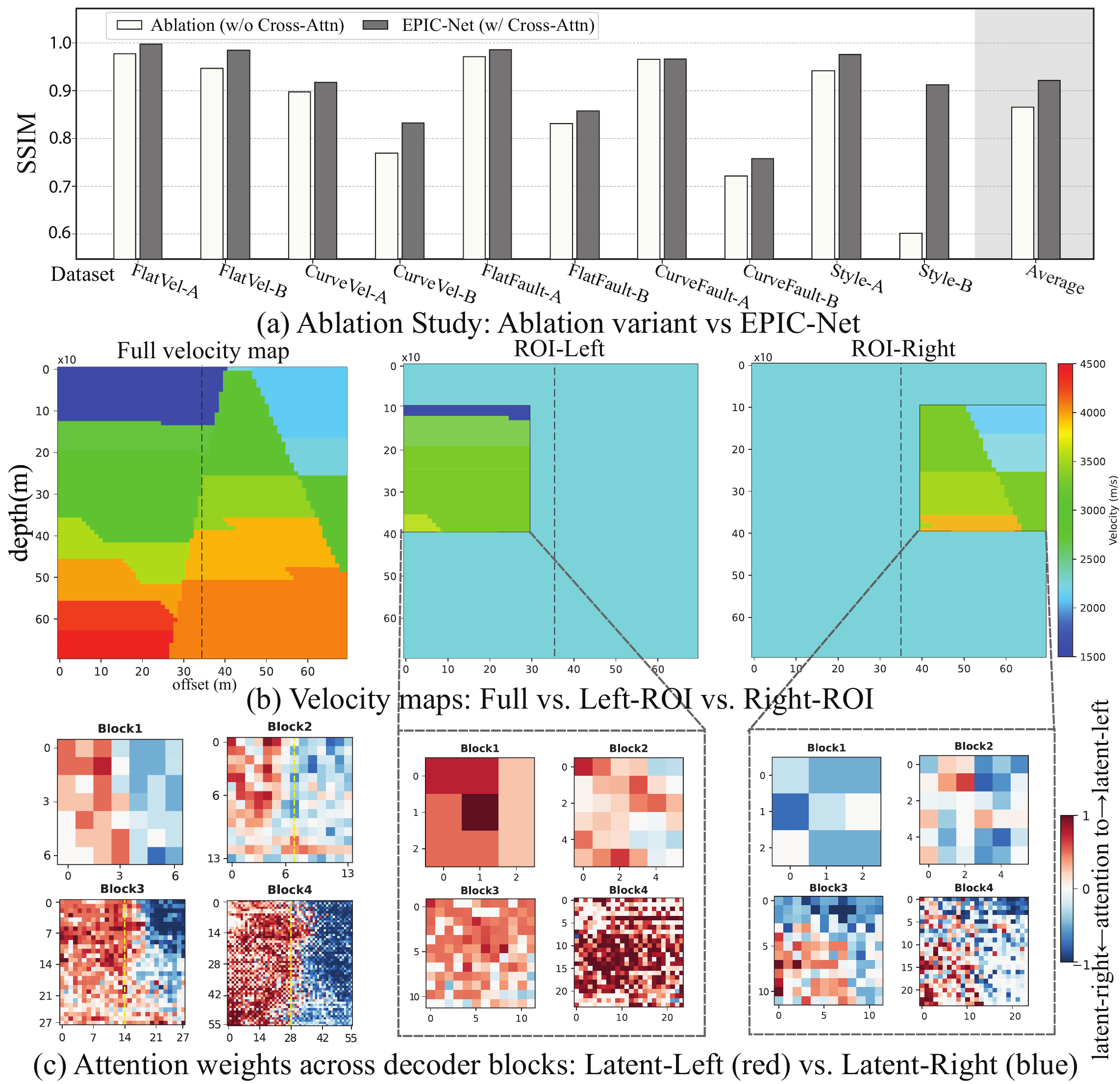}
\caption{Ablation study on the effectiveness of cross-attention: (a) ablation study w/ and w/o cross-attention; (b) reconstructed velocity maps and ROIs; (c) attention weight distribution across decoder blocks for two ROIs.}
\label{fig:attention}
\end{figure}

\textbf{Ablation study.}
To assess the contribution of the cross-attention module, Figure~\ref{fig:attention}(a) compares EPIC-Net with an ablated variant that removes cross-attention from the central decoder.
Across all ten datasets, EPIC-Net consistently achieves higher SSIM, with improvements ranging from moderate (e.g., FlatVel-A/B) to substantial on challenging datasets such as CurveVel-B and Style-B.
Removing cross-attention leads to noticeable degradation because the decoder can no longer fuse information across partitions, resulting in a loss of global wavefield coherence.
These results confirm the importance of physics-guidance in designing SciML.

\textbf{Physics-guided attention validation.}
Besides, we further visualize how EPIC-Net uses information from different spatial regions.
Figure \ref{fig:attention}(b)(c) shows the reconstructed velocity maps and the corresponding cross-attention weights for two ROIs.
For each ROI, we extract the attention maps across four decoder blocks.
Each heatmap indicates how much the decoder has higher weights for the latent on the left region (red) versus that on the right region (blue) when reconstructing each ROI.
The visualization reveals a clear spatial correspondence: the Left-ROI reconstruction primarily attends to latents from the left sensor group, while the Right-ROI reconstruction relies more heavily on latents from the right sensor group.
This behavior matches the physics of FWI, in which seismic waves recorded by nearby receivers contain the most relevant information for reconstructing the local subsurface.
These results confirm that EPIC-Net's cross-attention mechanism learns to fuse information in a physically meaningful way, naturally aligning attention patterns with the spatial distribution of seismic data.

\vspace{-5pt}
\section{Conclusion}

This paper presented EPIC, a distributed framework that enables practical field deployment of SciML by leveraging hardware and physics as two complementary teachers. 
Using full-waveform inversion as a representative SciML task, EPIC decomposes computation into lightweight edge encoding and physics-aware central decoding.
Evaluations are conducted on a testbed of five Raspberry Pi end devices under both Wi-Fi and emulated 4G conditions.
Results show that EPIC can reduce end-to-end latency by 8.9$\times$ and communication energy by 33.8$\times$ compared to centralized approaches, while achieving comparable or better reconstruction fidelity on 8 out of 10 OpenFWI datasets.
We demonstrate that hardware-physics co-guidance can produce superior distributed SciML systems, even outperforming centralized SciML.


\bibliographystyle{unsrt}
\bibliography{reference}

@article{wu2019inversionnet,
  title={InversionNet: An efficient and accurate data-driven full waveform inversion},
  author={Wu, Yue and Lin, Youzuo},
  journal={IEEE Transactions on Computational Imaging},
  volume={6},
  pages={419--433},
  year={2019},
  publisher={IEEE}
}

@inproceedings{deng2022openfwi,
  title={OpenFWI: Large-scale multi-structural benchmark datasets for full waveform inversion},
  author={Deng, Chengyuan and Feng, Shihang and Wang, Hanchen and Zhang, Xitong and Jin, Peng and Feng, Yinan and Zeng, Qili and Chen, Yinpeng and Lin, Youzuo},
  booktitle={Advances in Neural Information Processing Systems},
  volume={35},
  pages={6007--6020},
  year={2022}
}

@inproceedings{loshchilov2017decoupled,
  title={Decoupled Weight Decay Regularization},
  author={Loshchilov, Ilya and Hutter, Frank},
  booktitle={International Conference on Learning Representations},
  year={2019}
}

@article{SSIM,
  title={Image quality assessment: from error visibility to structural similarity},
  author={Wang, Zhou and Bovik, Alan C and Sheikh, Hamid R and Simoncelli, Eero P},
  journal={IEEE Transactions on Image Processing},
  volume={13},
  number={4},
  pages={600--612},
  year={2004},
  publisher={IEEE}
}

@article{jin2024empirical,
  title={An empirical study of large-scale data-driven full waveform inversion},
  author={Jin, Peng and Feng, Yinan and Feng, Shihang and Wang, Hanchen and Chen, Yinpeng and Consolvo, Benjamin and Liu, Zicheng and Lin, Youzuo},
  journal={Scientific Reports},
  volume={14},
  number={1},
  pages={20034},
  year={2024},
  publisher={Nature Publishing Group UK London}
}

@article{SciML1,
  title={Physics-informed neural networks: A deep learning framework for solving forward and inverse problems involving nonlinear partial differential equations},
  author={Raissi, Maziar and Perdikaris, Paris and Karniadakis, George E},
  journal={Journal of Computational Physics},
  volume={378},
  pages={686--707},
  year={2019},
  publisher={Elsevier}
}

@article{SciML2,
    author = {Sun, Jian and Innanen, Kristopher A. and Huang, Chao},
    title = {Physics-guided deep learning for seismic inversion with hybrid training and uncertainty analysis},
    journal = {Geophysics},
    volume = {86},
    number = {3},
    pages = {R303--R317},
    year = {2021},
    month = {03},
    issn = {0016-8033},
    doi = {10.1190/geo2020-0312.1},
    url = {https://doi.org/10.1190/geo2020-0312.1},
}

@ARTICLE{SciML3,
  author={Zhang, Jian and Sun, Hui and Zhang, Gan and Huang, Xingguo and Han, Li},
  journal={IEEE Transactions on Geoscience and Remote Sensing}, 
  title={Simultaneous Physics and Model-Guided Seismic Inversion Based on Deep Learning}, 
  year={2024},
  volume={62},
  number={},
  pages={1--11}
}

@ARTICLE{edge_computing1,
  author={Shi, Weisong and Cao, Jie and Zhang, Quan and Li, Youhuizi and Xu, Lanyu},
  journal={IEEE Internet of Things Journal}, 
  title={Edge Computing: Vision and Challenges}, 
  year={2016},
  volume={3},
  number={5},
  pages={637--646},
  doi={10.1109/JIOT.2016.2579198}}

@ARTICLE{edge_computing2,
  author={Satyanarayanan, Mahadev},
  journal={Computer}, 
  title={The Emergence of Edge Computing}, 
  year={2017},
  volume={50},
  number={1},
  pages={30--39},
  doi={10.1109/MC.2017.9}}

@article{geoscience1,
author = {J. Virieux and S. Operto},
title = {An overview of full-waveform inversion in exploration geophysics},
journal = {GEOPHYSICS},
volume = {74},
number = {6},
pages = {WCC1--WCC26},
year = {2009},
doi = {10.1190/1.3238367},

URL = {https://doi.org/10.1190/1.3238367},
}

@article{geoscience2,
author = {Albert Tarantola},
title = {Inversion of seismic reflection data in the acoustic approximation},
journal = {GEOPHYSICS},
volume = {49},
number = {8},
pages = {1259--1266},
year = {1984},
doi = {10.1190/1.1441754},
URL = {https://doi.org/10.1190/1.1441754}}

@ARTICLE{healthcare1,
  author={Matthews, Thomas P. and Wang, Kun and Li, Cuiping and Duric, Neb and Anastasio, Mark A.},
  journal={IEEE Transactions on Ultrasonics, Ferroelectrics, and Frequency Control}, 
  title={Regularized Dual Averaging Image Reconstruction for Full-Wave Ultrasound Computed Tomography}, 
  year={2017},
  volume={64},
  number={5},
  pages={811--825},
  doi={10.1109/TUFFC.2017.2682061}}

@inproceedings{FLA1,
  title={Communication-efficient learning of deep networks from decentralized data},
  author={McMahan, Brendan and Moore, Eider and Ramage, Daniel and Hampson, Seth and y Arcas, Blaise Aguera},
  booktitle={International Conference on Artificial Intelligence and Statistics},
  pages={1273--1282},
  year={2017},
  organization={PMLR}
}

@article{SLA1,
  title={Distributed learning of deep neural network over multiple agents},
  author={Gupta, Otkrist and Raskar, Ramesh},
  journal={Journal of Network and Computer Applications},
  volume={116},
  pages={1--8},
  year={2018},
  publisher={Elsevier}
}

@inproceedings{4G,
author = {Huang, Junxian and Qian, Feng and Gerber, Alexandre and Mao, Z. Morley and Sen, Subhabrata and Spatscheck, Oliver},
title = {A close examination of performance and power characteristics of 4G LTE networks},
year = {2012},
isbn = {9781450313018},
publisher = {Association for Computing Machinery},
address = {New York, NY, USA},
url = {https://doi.org/10.1145/2307636.2307658},
doi = {10.1145/2307636.2307658},
booktitle = {Proceedings of the 10th International Conference on Mobile Systems, Applications, and Services},
pages = {225--238},
numpages = {14},
location = {Low Wood Bay, Lake District, UK},
series = {MobiSys '12}
}

@inproceedings{Self_attention,
 author = {Vaswani, Ashish and Shazeer, Noam and Parmar, Niki and Uszkoreit, Jakob and Jones, Llion and Gomez, Aidan N and Kaiser, \L ukasz and Polosukhin, Illia},
 booktitle = {Advances in Neural Information Processing Systems},
 editor = {I. Guyon and U. Von Luxburg and S. Bengio and H. Wallach and R. Fergus and S. Vishwanathan and R. Garnett},
 pages = {},
 publisher = {Curran Associates, Inc.},
 title = {Attention is All you Need},
 url = {https://proceedings.neurips.cc/paper_files/paper/2017/file/3f5ee243547dee91fbd053c1c4a845aa-Paper.pdf},
 volume = {30},
 year = {2017}
}

@INPROCEEDINGS{Cross_attention,
  author={Lin, Hezheng and Cheng, Xing and Wu, Xiangyu and Shen, Dong},
  booktitle={2022 IEEE International Conference on Multimedia and Expo (ICME)}, 
  title={CAT: Cross Attention in Vision Transformer}, 
  year={2022},
  volume={},
  number={},
  pages={1--6},
  doi={10.1109/ICME52920.2022.9859720}}

@article{physical_laws1,
  title = {Enforcing Analytic Constraints in Neural Networks Emulating Physical Systems},
  author = {Beucler, Tom and Pritchard, Michael and Rasp, Stephan and Ott, Jordan and Baldi, Pierre and Gentine, Pierre},
  journal = {Physical Review Letters},
  volume = {126},
  issue = {9},
  pages = {098302},
  numpages = {7},
  year = {2021},
  month = {Mar},
  publisher = {American Physical Society},
  doi = {10.1103/PhysRevLett.126.098302},
  url = {https://link.aps.org/doi/10.1103/PhysRevLett.126.098302}
}

@article{physical_laws2,
author = {Willard, Jared and Jia, Xiaowei and Xu, Shaoming and Steinbach, Michael and Kumar, Vipin},
title = {Integrating Scientific Knowledge with Machine Learning for Engineering and Environmental Systems},
year = {2022},
issue_date = {April 2023},
publisher = {Association for Computing Machinery},
address = {New York, NY, USA},
volume = {55},
number = {4},
issn = {0360-0300},
url = {https://doi.org/10.1145/3514228},
journal = {ACM Computing Surveys},
month = nov,
articleno = {66},
numpages = {37}
}

@inproceedings{SciML4,
  title={A novel diffusion model for pairwise geoscience data generation with unbalanced training dataset},
  author={Yang, Junhuan and Zhang, Yuzhou and Sheng, Yi and Lin, Youzuo and Yang, Lei},
  booktitle={Proceedings of the AAAI Conference on Artificial Intelligence},
  volume={39},
  pages={21965--21973},
  year={2025}
}

@inproceedings{edge_computing3,
  title={On-device unsupervised image segmentation},
  author={Yang, Junhuan and Sheng, Yi and Zhang, Yuzhou and Jiang, Weiwen and Yang, Lei},
  booktitle={2023 60th ACM/IEEE Design Automation Conference (DAC)},
  pages={1--6},
  year={2023},
  organization={IEEE}
}

@article{edge_computing4,
  title={DiGiT: A Diffusion-based Modular Geophysical Toolkit for On-device Multi-modal Data Generation},
  author={Yang, Junhuan and Zhang, Yuzhou and Sheng, Yi and Lin, Youzuo and Jiang, Weiwen and Yang, Lei},
  journal={ACM Transactions on Embedded Computing Systems},
  year={2025},
  doi={10.1145/3779425},
  publisher={ACM}
}

@inproceedings{healthcare2,
  title={Toward fair ultrasound computing tomography: Challenges, solutions and outlook},
  author={Sheng, Yi and Yang, Junhuan and Lin, Youzuo and Jiang, Weiwen and Yang, Lei},
  booktitle={Proceedings of the Great Lakes Symposium on VLSI 2024},
  pages={748--753},
  year={2024}
}

@inproceedings{healthcare3,
  title={Enhanced ai for science using diffusion-based generative ai-a case study on ultrasound computing tomography},
  author={Yang, Junhuan and Sheng, Yi and Zhang, Yuzhou and Wang, Hanchen and Lin, Youzuo and Yang, Lei},
  booktitle={Proceedings of the Great Lakes Symposium on VLSI 2024},
  pages={754--759},
  year={2024}
}

@inproceedings{physical_laws3,
  title={Edgeo: A physics-guided generative ai toolkit for geophysical monitoring on edge devices},
  author={Yang, Junhuan and Wang, Hanchen and Sheng, Yi and Lin, Youzuo and Yang, Lei},
  booktitle={Proceedings of the 61st ACM/IEEE Design Automation Conference (DAC)},
  pages={1--6},
  year={2024}
}

\end{document}